# A New Statistical Framework for Genetic Pleiotropic Analysis of High Dimensional Phenotype Data

Panpan Wang,[2,1] Mohammad Rahman,[1] Li Jin[2,**] and Momiao Xiong[1,*]

[1]Human Genetics Center, Department of Biostatistics, The University of Texas School of Public Health, Houston, TX 77030, USA

[2]State Key Laboratory of Genetic Engineering and Ministry of Education, Key Laboratory of Contemporary Anthropology, Collaborative Innovation Center for Genetics and Development, School of Life Sciences and Institutes of Biomedical Sciences, Fudan University, Shanghai, 200433, China

**Running Title**: Genotype-Phenotype Network Analysis

[*]Address for correspondence and reprints: Dr. Momiao Xiong, Human Genetics Center, The University of Texas Health Science Center at Houston, P.O. Box 20186, Houston, Texas 77225, (Phone): 713-500-9894, (Fax): 713-500-0900, E-mail: Momiao.Xiong@uth.tmc.edu

[**] Address for correspondence and reprints: Dr. Li Jin, State Key Laboratory of Genetic Engineering and Ministry of Education, Key Laboratory of Contemporary Anthropology, Collaborative Innovation Center for Genetics and Development, School of Life Sciences and Institutes of Biomedical Sciences, Fudan University, Shanghai 200433, China, E-mail: lijin.fudan@gmail.com

## Abstract

The widely used genetic pleiotropic analysis of multiple phenotypes are often designed for examining the relationship between common variants and a few phenotypes. They are not suited for both high dimensional phenotypes and high dimensional genotype (next-generation sequencing) data. To overcome these limitations, we develop sparse structural equation models (SEMs) as a general framework for a new paradigm of genetic analysis of multiple phenotypes. To incorporate both common and rare variants into the analysis, we extend the traditional multivariate SEMs to sparse functional SEMs. To deal with high dimensional phenotype and genotype data, we employ functional data analysis and the alternative direction methods of multiplier (ADMM) techniques to reduce data dimension and improve computational efficiency. Using large scale simulations we showed that the proposed methods have higher power to detect true causal genetic pleiotropic structure than other existing methods. Simulations also demonstrate that the gene-based pleiotropic analysis has higher power than the single variant-based pleiotropic analysis. The proposed method is applied to exome sequence data from the NHLBI's Exome Sequencing Project (ESP) with 11 phenotypes, which identifies a network with 137 genes connected to 11 phenotypes and 341 edges. Among them, 114 genes showed pleiotropic genetic effects and 45 genes were reported to be associated with phenotypes in the analysis or other cardiovascular disease (CVD) related phenotypes in the literature.

## Introduction

In the past several years, a large number of statistical methods for association analysis of both qualitative and quantitative traits with next-generation sequencing data were developed.[1-14] Most genetic analyses of quantitative traits focus on association analysis of a single trait, analyzing each phenotype individually and independently.[15] However, multiple phenotypes are correlated. For example, metabolism of lipoproteins involves cholesterol, triglycerides, very low density lipoproteins (VLDL), low density lipoproteins and high density lipoproteins. These multiple traits are dependent. The integrative analysis of correlated phenotypes often increase statistical power to identify genetic associations.[16,17] The association analysis of multiple phenotypes is expected to become popular in the near future.[18]

Three major approaches are commonly used to explore association of genetic variants with multiple correlated phenotypes: multiple regression methods, integration of p values of univariate analysis, and dimension reduction methods.[16] Despite their differences in selection of specific methods for estimation, all these estimation methods share the following common features. First, many methods were designed for common variants and hence may not be appropriate for rare ones. Second, the results of all these analyses are difficult to interpret. They do not provide information to indicate which phenotypes the genetic variants are significantly associated.[15] Third, all these methods estimate the effect of the genetic variant on each phenotype individually and do not explore the dependency patterns of genetic effects among the phenotypes and do not provide a detailed characterization of the relationships among the genetic effects. Fourth, all these estimations only estimate the effects of the genetic variants on the phenotypes. However, the genetic effects can be classified into three types of effects: direct, indirect and total effects. These methods are unable to reveal mechanisms underlying the genetic structures of multiple phenotype association analysis.[19] The direct effect is the measurement of the influence of a genetic variant on a phenotype that is not mediated by any other phenotypes in a system. The indirect effect of a genetic variant measures the sensitivity of a phenotype to change of a genetic variant that is mediated by at least one intervening variable (phenotype). The total effect is the sum of the direct and indirect effects. The most popular multivariate association methods are lack of ability to decompose total effect into direct effect and indirect effect and ignore indirect effects through other mediating phenotypes and risk factors. Therefore, they cannot discover how the effect of the genetic variant on the phenotype is mediated by other phenotypes and the effect path from the initially affected phenotype by the genetic variant through a number of mediating phenotypes to the targeted phenotype. Pleiotropic effect is a context dependent genetic effect and plays an important role in multivariate trait association studies and evolution analysis.[20] The pleiotropic effect of a specific genetic variant on multiple phenotypes may be due to either direct contribution of the genetic variant to the multiple phenotypes or phenotype correlations (mediations). The multivariate trait association studies cannot distinguish the paths connecting multiple phenotypes and genetic effects.[21]

In the past several years, there have been increasing interests in modeling the complex structures among phenotypes, risk factors and genotypes which are referred to as the genotype-phenotype networks and therefore overcome these limitations. Current methods for inference of genotype-phenotype networks can be classified into two categories: whole network scoring methods and local analysis methods.[22-29] Network scoring approaches assign a score to the network model for measuring how well the network fits the data and develop algorithms to search the network with the best score. Local analysis methods analyze small sets of variables that are pieced together into networks from multiple causality tests between variables.

One of network scoring methods is structure equations that can be used as a tool to model the complex network structures among phenotypes, risk factors and genotypes.[19-21,30-32] A graphical model in which the variables are represented as nodes and the relationships between variables are represented by edges between the nodes can be used to model the genotype-phenotype networks. Structural equations can generate biological interpretations of relations among variables and uncover the mechanism structure underlying phenotypic and genotypic relationships. To date, in applications of the structural equation model (SEM) in quantitative genetics, the causal structure was assumed to be known as a priori, or partially specified, thereby allowing selection of the causal structure for a small set of variables from the data.[21] There are two major approaches to estimate the causal structure from the data. One approach is based on the conditional independence and the notion of Markov equivalence of directed acyclic graphs (DAGs).[33] DAGs encode causal structure. However, a DAG is not, in general, identifiable from observational data. Conditional independence only determines the skeleton of the DAG which is the undirected graph of the DAG by removing its directions of all edges, and the $v$ structure of the DAG where two nodes are directed to a common node (collider).[34] A number of algorithms such as PC-algorithms have been used to estimate the equivalence class of DAGs.[35] A second approach is to use the notion of 'sparse' and develop sparse SEMs for estimating the causal structures.[36] By incorporating the penalized constraints of the parameters into the likelihood function to enforce the network sparsity, we could estimate the causal structure. Coordinate ascent algorithms are often used to maximize the penalized likelihood functions.

Despite their successful application to joint analysis of genetic architecture and causal phenotype networks, current approaches often demand intensive computations and are lack of efficient computational algorithms for implementing penalization of network structure parameters. Therefore, they cannot be used for large-scale causal inference. Most current approaches are designed for common variants and are difficult to be applied to NGS data. The purpose of this paper is to overcome these limitations. We first develop novel functional SEMs where exogenous genotype profiles across a genomic region or a gene are represented as a function of the genomic position for genetic association analysis of multiple quantitative traits which is referred to as multivariate QTL analysis. The functional SEMs for multivariate QTL analysis consist of three components. The first component is a phenotype network that is modeled as a directed graph. The second component is a genotype network that is represented as an undirected graph. The third component is connections between the genotype network and phenotype network with direction from genotype nodes to phenotype nodes. To make the network sparse and reduce the burden of computations, we develop the novel sparse SEMs for genotype-phenotype networks and an efficient computational algorithms based on alternative direction methods of multiplier (ADMM) to search the causal structure and estimate the parameters.[37,38] We will estimate the direct, indirect and total effects of the genetic variants on the phenotypes using estimated directed graph and intervention calculus[39] and explore the relationships between direct, indirect and total effects estimated from SEMs and the genetic effects estimated from the traditional simple regressions and multiple regressions. Finally, the sparse SEMs are applied to exome sequence data from the NHLBI's Exome Sequencing Project (ESP) with 11 phenotypes. A program for implementing the developed sparse SEMs for quantitative genetic analysis with multiple phenotypes can be downloaded from our website http://www.sph.uth.tmc.edu/hgc/faculty/xiong/index.htm.

## Material and Methods

Multivariate quantitative trait association analysis can be investigated by phenotype-genotype networks, which can be represented as a graph. Phenotypes, covariates such as age, sex, race, and SNPs are variables. Variables are represented as nodes in the graph. We assume that causal relationships among phenotypes exist. Therefore, a phenotype network is represented by a directed graph. A directed edge between two nodes indicates the causal relationship between them. Since SNPs do not have causal relationships among them, a genotype network is represented as an undirected graph. An edge between two nodes in the genotype network indicates their correlation. Since all SNPs and covariates may cause changes in phenotypes, the phenotype network and genotype network are connected by edges directed from covariates and SNP to the phenotypes. The phenotypes and connections between phenotypes, covariates and SNPs can be modeled by structural equations. The genotype network can be leant by graphical LASSO (GLASSO),[39] here we didn't focus on genotype network in this paper. An example of phenotype-genotype network is shown in Supplementary Figure S1.

### *SEMs for Multivariate Association Analysis*

The SEMs offer a general statistical framework for inferring phenotype networks and connections between genotypes and phenotypes. Assume that $n$ individuals are sampled. We consider $M$ phenotypes that are referred to as endogenous variables. We denote the $n$ observations on the $M$ endogenous variables by the matrix $Y = [y_1, y_2, ..., y_M]$, where $y_i = [y_{1i}, ..., y_{ni}]^T$ is a vector of collecting $n$ observation of the endogenous variable $i$. Covariates, genetic variants as exogenous or predetermined variables are denoted by $X = [x_1, ..., x_K]$ where $x_i = [x_{1i,}, ..., x_{ni}]^T$ .Similarly, random errors are denoted by $E = [e_1, ..., e_M]$, where we assume $E[e_i]=0$ and $E[e_i e_i^T]=\sigma_i^2 I_n$ for $i=1,...,M$. The linear structural equations for modeling relationships among phenotypes and genotypes can be written as[38]

$$\begin{aligned} &y_1\gamma_{11}+y_2\gamma_{21}+...+y_M\gamma_{M1}+x_1\beta_{11}+x_2\beta_{21}+...+x_K\beta_{K1}+e_1=0 \\ &\vdots \\ &y_1\gamma_{1M}+y_2\gamma_{2M}+...+y_M\gamma_{MM}+x_1\beta_{1M}+x_2\beta_{2M}+...+x_K\beta_{KM}+e_M=0 \end{aligned} \tag{1}$$

where the $\gamma$ 's and $\beta$ 's are the structural parameters of the system that are unknown. In matrix notation the SEMs (1) can be rewritten as

$$Y\Gamma+XB+E=0, \tag{2}$$

where $\Gamma=[\Gamma_1,...,\Gamma_M]$, $\Gamma_i=[\gamma_{1i},...,\gamma_{Mi}]^T$, $B=[B_1,...,B_M]$, $B_i=[\beta_{1i},...,\beta_{Ki}]^T$.

We assume that the random errors in the structural equations are independent and uncorrelated with exogenous variables. We apply the sparsity penalty to each equation to ensure that the sparse SEMs are identifiable.

***Two-Stage Least Square Estimates of the Parameters in the SEMs***

The ordinary least squares estimator is biased and inconsistent for the parameters of structural equations. To ensure the consistent estimates of the parameters in the SEMs, we use a generalized least square method that can be interpreted as a two-stage least square estimate method to estimate the parameters in the SEMs.[38]

Recalling that $y_i$ is the vector of observations of the variable $i$, let $Y_{-i}$ be the observation matrix $Y$ after removing $y_i$ from it and $\gamma_{-i}$ be the parameter vector $\Gamma_i$ after removing the parameter $\gamma_{ii}$. The $i$ th equation:

$$Y\Gamma_i+XB_i+e_i=0$$

can be rewritten as

$$\begin{aligned} y_i &= Y_{-i}\gamma_{-i} + XB_i + e_i \\ &= W_i\Delta_i + e_i , \end{aligned} \tag{3}$$

where $W_i = \begin{bmatrix} Y_{-i} & X \end{bmatrix}, \Delta_i = \begin{bmatrix} \gamma_{-i}^T & B_i^T \end{bmatrix}^T$.

Multiplying by the matrix $X^T$ on both sides of equation (3), we obtain

$$X^T y_i = X^T Y_{-i}\gamma_{-i} + (X^T X)B_i + X^T e_i = X^T W_i \Delta_i + X^T e_i . \tag{4}$$

It is known that

$$\mathrm{cov}(X^T e_i, X^T e_i) = X^T X\sigma_i^2 .$$

The generalized least square estimate $\hat{\Delta}_i$ is given by

$$\hat{\Delta}_i = [W_i^T X(X^T X)^{-1} X^T W_i]^{-1} W_i^T X(X^T X)^{-1} X^T y_i . \tag{5}$$

The generalized least square estimate $\hat{\Delta}_i$ can be interpreted as a two-stage least square estimate.[38]

Suppose that in the first stage, $Y_{-i}$ is regressed on $X$ to obtain

$$\hat{\Pi}_i = (X^T X)^{-1} X^T Y_{-i} \text{ and } \hat{Y}_{-i} = X\hat{\Pi}_i .$$

Then,

$$\begin{aligned} \hat{W}_i &= \begin{bmatrix} \hat{Y}_{-i} & X \end{bmatrix} \\ &= X(X^T X)^{-1} X^T W_i . \end{aligned}$$

Equation (5) can be reduced to

$$\hat{\Delta}_i = (\hat{W}_i^T \hat{W}_i)^{-1} \hat{W}_i^T y_i \, . \qquad (6)$$

Therefore, if $W_i$ in equation (3) is replaced by $\hat{W}_i$, equation (6) can be interpreted as that in the second stage, $y_i$ is regressed on $\hat{Y}_i$ and $X$ to obtain estimate $\hat{\Delta}_i$.

***Sparse SEMs and Alternating Direction Method of Multipliers***

In general, the genotype-phenotype networks are sparse. Therefore, $\Gamma$ and $B$ are sparse matrices. In order to obtain sparse estimates of $\Gamma$ and $B$, the natural approach is the $l_1$-norm penalized regression of equation (4). Using weighted least square and $l_1$-norm penalization, we can form the following optimization problem:

$$\begin{aligned} &\min_{\Delta_i} \quad f(\Delta_i) + \lambda \| \Delta_i \|_1 \\ &\text{where } f(\Delta_i) = (X^T y_i - X^T W_i \Delta_i)^T (X^T X)^{-1} (X^T y_i - X^T W_i \Delta_i). \end{aligned} \qquad (7)$$

The size of the genotype-phenotype network may be large. An efficient alternating direction method of multipliers (ADMM)[37] is used to solve the optimization problem (7). The procedure for implementing ADMM is given below (more detailed descriptions are provided in Appendix A).

Algorithm:

For $i = 1, \ldots, M$

Step 1. Initialization

$$\begin{aligned} &u^0 := 0 \\ &\Delta_i^0 := [W_i^T X (X^T X)^{-1} X^T W_i + \rho I]^{-1} W_i^T X (X^T X)^{-1} X^T y_i \\ &Z_i^0 := \Delta_i^0. \end{aligned}$$

Carry out steps 2,3 and 4 until convergence

Step 2.

$$\Delta_i^{(k+1)} := [\frac{1}{\rho} I - \frac{1}{\rho} W_i^T X(\rho X^T X + X^T W_i W_i^T X)^{-1} X^T W_i][W_i^T X(X^T X)^{-1} X^T y_i + \rho(Z_i^k - u^k)]$$

Step 3.

$$Z_i^{(k+1)} := \text{sgn}(\Delta_i^{k+1} + u^k)(|\Delta_i^{k+1} + u^k| - \frac{\lambda}{\rho})_+,$$

where

$$|x|_+ = \begin{cases} x & x \geq 0 \\ 0 & x < 0. \end{cases}$$

Step 4.

$$u^{(k+)} :== u^{(k)} + (\Delta_i^{(k+1)} - Z_i^{(k+1)}).$$

Under some assumptions convergence of ADMM can be proved.[37] In practice, although it can be slow to converge to high accuracy, ADMM converges to modest accuracy within a few tens of iterations. When large-scale problems and parameter estimation problems are considered, modest accuracy is sufficient. Therefore, ADMM may work very well for structure and parameter estimation in the genotype-phenotype networks.

Most of the elements of matrices $\Gamma$ and $B$ are equal to zero. The $l_1$ – regularized Lasso for the two stage least squares approach and ADMM algorithms are expected to shrink most of the coefficient matrices $\Gamma$ and $B$ toward zero, yielding sparse network structures. The sparsity-controlling parameter $\lambda$ will be estimated via cross validation or set by users to get reasonable results.

***Sparse Functional Structural Equation Models for Phenotype and Genotype Networks***

In the previous section, the SEMs carry out variant by variant analysis. However, the power of the traditional variant-by-variant analytical tools that are mainly designed for common variants, for association analysis of rare variants with the phenotypes will be limited. Large simulations have shown that combining information across multiple variants in a genomic region of analysis will greatly enhance power to detect association of rare variants.[9] To utilize multi-locus genetic information, we propose to use a genomic region or a gene as a unit in multiple trait association analysis and develop sparse functional structural equation models (FSEMs) for construction and analysis of the phenotype and genotype networks.

Let $t$ be a genomic position. Define a genotype profile $x_i(t)$ of the $i$-th individual as

$$x_i(t) = \begin{cases} 2P_q(\mathrm{t}), & \mathrm{QQ} \\ P_q(\mathrm{t}) - \mathrm{P_Q(t)}, & \mathrm{Qq} \\ -2P_Q(\mathrm{t}), & \mathrm{qq} \end{cases}$$

where $Q$ and $q$ are two alleles of the marker at the genomic position $t$, $P_Q(t)$ and $P_q(t)$ are the frequencies of the alleles $Q$ and $q$, respectively. Suppose that we are interested in $k$ genomic regions or genes [$a_j, b_j$], denoted as $T_j, j = 1,2,...,k$. We consider the following functional structural equation models:

$$
\begin{aligned}
&y_1\gamma_{11} + y_2\gamma_{21} + ... + y_M\gamma_{M1} + \int_{T_1} x_1(t)\beta_{11}(t)dt + ... + \int_{T_k} x_k(t)\beta_{k1}(t)dt + e_1 = 0 \\
&y_1\gamma_{12} + y_2\gamma_{22} + ... + y_M\gamma_{M2} + \int_{T_1} x_1(t)\beta_{12}(t)dt + ... + \int_{T_k} x_k(t)\beta_{k2}(t)dt + e_2 = 0 \\
&\vdots \qquad\qquad\qquad\qquad \vdots \qquad\qquad\qquad\qquad \vdots \\
&y_1\gamma_{1M} + y_2\gamma_{2M} + ... + y_M\gamma_{MM} + \int_{T_1} x_1(t)\beta_{1M}(t)dt + ... + \int_{T_k} x_k(t)\beta_{kM}(t)dt + e_M = 0
\end{aligned} \tag{8}
$$

where $\beta_{ij}(t)$ are genetic effect functions.

For each genomic region or gene, we use functional principal component analysis to calculate principal component function.[14] We expand $x_{nj}(t), n = 1,...,N, j = 1,2,...,k$ in each genomic region in terms of orthogonal principal component functions:

$$x_{ij}(t) = \sum_{l=1}^{L_j} \eta_{ijl}\phi_{jl}(t), \; j = 1,...,k,$$

where $\phi_{jl}(t), j = 1,...,k, l = 1,...,L_j$ are the $l$-th principal component function in the $j$-th genomic region or gene and $\eta_{ijl}$ are the functional principal component scores of the $i$-th individual.

Let $\eta$ be a matrix collection of all functional principal component scores, the parameter matrix $B$ can be defined as that in Appendix B, matrices $Y$ and $\Gamma$ can be defined as that in the previous section. The structural functional equations can be reduced in terms of functional principal component scores (Appendix B):

$$Y\Gamma_i + \eta B_i + e_i = 0,$$

which can be rewritten as

$$y_i = W_i \Delta_i + e_i,$$

where $W_i = [Y_{-i} \quad \eta], \Delta_i = [\gamma_{-i}^T \quad B_i^T]^T$.

Then, the sparse FSEMs are transformed to

$$\min_{\Delta_i} \quad f(\Delta_i) + \lambda \| \Delta_i \|_1$$
$$\text{where } f(\Delta_i) = (\eta^T y_i - \eta^T W_i \Delta_i)^T (\eta^T \eta)^{-1} (\eta^T y_i - \eta^T W_i \Delta_i). \tag{9}$$

The ADMM algorithms for solving the sparse FSEMs are the same as that in the previous section if the matrix $X$ is replaced by a functional principal component score matrix $\eta$ (Appendix B).

***Effect Decomposition and Estimation***

To make this paper self-contained, we introduce basic concepts and methods for decomposition and estimation of the effects. In the genotype-phenotype network analysis we are interested in estimation of effects of genetic variants on phenotypes, which is referred to as genetic effects and effects of treatment on phenotypes. All genetic effects and treatment effects can be decomposed as total (causal), direct effects and indirect effects. Distinction between total, direct and indirect effects are of great practical importance in genetic association analysis.[40] The total effect measures the changes of response variable $Y$ (phenotype) would take on the value $y$ when variable $X$ is set to $x$ by external intervention. Direct effect is defined as sensitivity of $Y$ to changes in $X$ while all other variables in the model are held fixed. Indirect effect is to measure the portion of the effect which can be explained by mediation alone, while inhibiting the capacity of $Y$ to respond to $X$.[41] The total effect is equal to the summation of direct and indirect effects.

Given a directed graph model $G$, one can compute total effects using intervention calculus.[34,42] Suppose that the expected value of a response variable $Y$, after $X$ is assigned value $x$ by intervention is denoted by $E[Y \mid do(X = x)]$. The total effect is defined as

$$\frac{\partial}{\partial x} E[Y \mid do(X = x)]. \tag{10}$$

Note $X_j$ is called a parent of $X$ in $G$ if there is a directed edge $X_j \to X$. Let $\mathrm{pa}_x$ denote the set of all parents of $X$ in $G$. In the linear SEMs, we assume that $E[Y \mid X, \mathrm{pa}_x]$ is linear in $X$ and $\mathrm{pa}_x$:

$$E[Y \mid X, \mathrm{pa}_x] = \alpha + \beta X + \gamma^T \mathrm{pa}_x . \qquad (11)$$

Then,

$$\frac{\partial}{\partial x} E[Y \mid do(X = x)] = \beta .$$

When a directed graph is given, it is easy to calculate total effect.[42] Assume that there are $k$ directed paths from $X$ to $Y$ and $p_i$ are the product of the path coefficients along the $i$-th path. The total effect of $X$ on $Y$ is then defined as $\sum_{i=1}^{k} p_i$. As shown in Figure S2, the total effect of $X$ on $Y$ is $ag + bdh + acdh$. By its definition, direct effect measures the sensitivity of $Y$ to changes in $X$ while all other variables in the model are held fixed. In other words, all links from $X$ to $Y$ other than the direct link will be blocked. As a consequences, the direct effect is equal to the path coefficient from $X$ to $Y$. In the linear SEMs, the indirect effect of $X$ on $Y$ mediated by $M$ is equal to the sum of the products associated with directed paths from $X$ to $Y$ through $M$.[42] In Figure S2, there is no direct effect from $X$ to $Y$. The indirect effect of $X$ on $Y$ which is mediated by $B$ and $D$ is equal to $bdh$.

In the SEMs for genotype-phenotype networks, since all SNPs only form undirected graph and there are no directed links between SNPs although we can observe linkage (or correlation) between SNPs; SNPs in the genotype-phenotype networks do not have parents. The total effect of SNP $X$ on $Y$ is the regression coefficient $\beta$ of the following linear regression:

$$E[Y \mid do\,(X = x)] = \alpha + \beta x\,,$$

which is a simple regression of $Y$ on $X$. This indicates that the traditional simple regression for association studies captures the total effect of a genetic variant on a phenotype.

If we include environments and risk factors such as smoking and obesity in the model and want to evaluate the effects of the environments and risk factors on the phenotype, these variables play mediating roles and will also be taken as phenotypes. We denote these mediating phenotypes by $Y_{ME}$. Since genetic variants, and other risk factors and phenotypes will affect the mediating phenotypes, the mediating phenotypes in the graphics may have parents. Their parents are denoted by $S$. Total effect of the mediation phenotype on the target phenotype is calculated by

$$E[Y \mid do\,(Y_{ME} = y_{ME}, X_{pa} = x_{pa})] = \alpha + \beta y_{ME} + \gamma^T x_{pa}\,, \qquad (12)$$

where $\beta$ is the total effect of the mediation phenotype $Y_{ME}$ on the target phenotype $Y$. In this case, a simple regression of $Y$ on $Y_{ME}$ can no longer be used to measure the total effect of the mediation phenotype $Y_{ME}$ on the target phenotype $Y$. To observe this, we simulated 1,000 individuals with the SEM as shown in Figure S3. Each variable has a noise term distributed as $N(0,1)$. The total effect of the mediation phenotype $Y_{ME}$ on the target phenotype $Y$ is 3.5. We obtain the simple regression:

$$Y = 1.39 + 5.85 Y_{ME}\,.$$

It is clear that the coefficient of the simple regression is 5.85. This value is far away from the total effect 3.5. However, using equation (12) we obtain

$$Y = 3.54 Y_{ME} + 5.85 X ,$$

where the regression coefficient 3.54 measured the total effect of the mediation phenotype $Y_{ME}$ on the target phenotype $Y$ .

***Test Statistics for Path Coefficients***

Testing connection between the $j$ -th gene and the $i$ -th phenotype in the genotype-phenotype network, we formally investigate the problem of testing the coefficient of the path directed from the $j$ - th gene to the $i$ -th phenotype:

$$H_0 : \beta_{ji}(t) = 0 \ , \qquad \forall t \in [0, T_j] \tag{13}$$

against

$$H_a : \beta_{ji}(t) \neq 0 \ .$$

If the coefficient function of path or genetic effect function $\beta_{ji}(t)$ is expanded in terms of the principal component functions:

$$\beta_{ji}(t) = \sum_{g=1}^{G} b_{jig} \phi_{jg}(t) ,$$

then testing the null hypothesis $H_0$ in equation (13) is equivalent to testing the hypothesis:

$$H_0 : b_{jig} = 0 , \forall \mathrm{g} . \tag{14}$$

The path coefficients $b_{jig}$ can be estimated by solving problems (8) and (9). Let

$\hat{b}_{ji} = [b_{ji1}, ..., b_{jiG}]^T$ . The covariance matrix of the vector of the estimators of path coefficients for the $i$ -th equation is given by[38]

$$\hat{\Sigma}_i = \sigma_{ii} [W_i^T \eta (\eta^T \eta)^{-1} \eta^T W_i]^{-1} , \tag{15}$$

where

$$\sigma_{ii} = (y_i - W_i\hat{\Delta}_i)^T (y_i - W_i\hat{\Delta}_i)/n \,. \tag{16}$$

Let $\Lambda_i$ be the submatrix that corresponds to $b_{ji}$ in the matrix $\hat{\Sigma}_i$. Define the statistic for testing the directed connection from the $j$-th gene to the $i$-th phenotype as

$$T_g = \hat{b}_{ji}^T \Lambda_i^{-1} \hat{b}_{ji} \,. \tag{17}$$

Under the null hypothesis of no association $H_0 : b_{ji} = 0$, $T_g$ is asymptotically distributed as a central $\chi^2_{(G)}$ distribution where $G$ is the number of functional principal components in the expansion of $\beta_{ji}(t)$.

For testing a single parameter or single variant's path coefficient in the SEMs, the $l$-th parameter of the $i$-th equation, the statistic is given by

$$T_c = \frac{\hat{\Delta}_{il}^2}{\operatorname{var}(\hat{\Delta}_{il})} \,, \tag{18}$$

where $\operatorname{var}(\hat{\Delta}_{il})$ is the $l$-th diagonal element of the matrix $\hat{\Sigma}_i$. Under the null hypothesis $H_0 : \Delta_{il} = 0$, $T_c$ is asymptotically distributed as a central $\chi^2_{(1)}$ distribution.

## Results

### *Model Evaluation by Simulations*

We evaluated the performance of the sparse SEM approach for genetic analysis of multiple quantitative traits in simulation studies of a genotype-phenotype network where SNP-based simulations and gene-based simulations were considered. The simulations were carried out for common variants, rare variants and both common and rare variants. In the SNP-based simulations, the SNPs consisted of 30 common variants, 30 rare variants and 10 common and 20 rare variants, respectively. In the gene-based simulations, a total of 10 genes (10 SNPs for each gene) were included where we also considered 3 scenarios: genes consisted of 1) only common variants, 2) only rare variants and 3) both rare and common variants. The genotype data were selected from the NHLBI's Exome Sequencing Project (ESP) with 3,248 individuals of European origin, which were then used to generate a population of 1,000,000 individuals.

We first study the SNP-based simulations. The genotype-phenotype network consisted of two parts. The first part was the phenotype network that was modeled by a DAG. The second part was the connections between the genotypes and phenotypes in which the genotypes were directed to the phenotypes. We randomly generated a genotype-phenotype network structure (Figure S4). The parameters $\Gamma_{ij}$ in the SEMs for modeling phenotype sub-network were generated from a uniformly distributed random variable over the interval (0.5, 1) or (-1,-0.5) if an edge from node $j$ to node $i$ was presented in the phenotype sub-network; otherwise $\Gamma_{ij}$=0. Similarly, the parameters $B_{ij}$ in the SEMs for modeling the direction from the genotype (SNP) node $j$ to the phenotype node $i$ were generated from a uniformly distributed random variable over the interval (0, 1) or (-1,0) if an edge from node $j$ to node $i$ was presented in the genotype-phenotype network, otherwise $B_{ij} = 0$. The indicator variables for coding genotypes of the SNP were as previously described. The true structure of the genotype-phenotype network used for simulation was plotted in Figure S4 where no cycles were presented. Using the pre-determined network structure, the assumed parameters in the structural equations, we generated the phenotypes by the model: $Y = -XB\Gamma^{-1} + \varepsilon\Gamma^{-1}$, where $\varepsilon \sim N(0, 0.01 \times I)$, and $X$ is a matrix of indicator variables for coding genotypes.

Simulations were repeated 1,000 times. Five-fold cross validation was used to determine the penalty parameter $\lambda$ that was then employed to infer the network while running power simulations. Two measures: the power of detection (PD) and the false discovery rate (FDR) were used to evaluate the performance of the algorithms for identification of the network structures. Specifically, let $N_t$ be the total number of edges among 1000 replicates of the network and $\hat{N}_t$ be the total number of edges detected by the inference algorithm, $N_{true}$ be the total number of true edges detected among simulated network and $N_{False}$ be the false edges detected among $\hat{N}_t$. Now, the power of detection (PD) is defined by $\frac{N_{True}}{\hat{N}_t}$ and false discovery rate (FDR) is defined by $\frac{N_{False}}{\hat{N}_t}$.

In the SNP-based simulations we compared the performance of the proposed sparse 2-stage SEM (S2SEM) method with the QTLnet algorithm[22] which can be used for joint inference of causal network and genetic architecture for correlated phenotypes. Figure 1 shows the power of two methods: S2SEM and QTLnet for detecting the structure of the genotype (common variants, rare variants and both common and rare variants)– phenotype network as a function of sample size, assuming the network sparsity level $= 0.008$. We observed three features. First, when the network was sparse (the sparsity level $= 0.008$) the power of S2SEM in all three cases was the highest among three methods. Second, the power of the two methods to detect the structure of the networks with the common variants was the highest, followed by the both common and rare variants. The power of two methods to detect the structure of the network with the rare variants was the lowest. Third, in general, the power increased when the sample sizes increased. To fully evaluate the performance of the two methods, we also presented Figure 2 showing the FDR for detection of the structure of the networks as a function of sample sizes. It was clear that the FDR of the S2SEM in all three cases was lower than QTLnet method. The FDR of two methods to detect the structure of the networks with the common variants was the lowest, followed by the both common and rare variants. The FDR of two methods to detect the structure of the network with the rare variants was the highest. However, the false discovery rates for these two methods and in three cases were larger than 0.1 even the sample sizes reached 3,000, and it is larger than 0.3 in the rare variants case.

Figure 1 showed that the power of the variant by variant tests for identifying the network structure with the rare variants was low. To increase the power, we develop functional SEMs (FSEMs) for network analysis using a genomic region or gene as a unit of analysis. To evaluate this strategy, we present Figures 3 and 4 to compare the power and FDR of the gene-based FSEMs and the SNP-based SEMs for detection of the network structures. Since the original papers for QTLnet[22] did not develop the gene-based statistics, in Figures 3 and 4 we did not present the results of QTLnet algorithm. We observed that in all three cases: common, rare and both common and rare variants, the gene-based FSEM had much higher power and smaller FDR than the SNP-based SEMs. It is interesting to observe that even if for the rare variants the gene-based method can reach the power as high as 85% when sample sizes were larger than 3,000.

***Application to Real Data Examples***

To evaluate its performance, we applied the sparse functional SEMs with a gene as a unit of analysis to a sample of 1,011 European-Americans (EA) with complete exome sequencing (total of 1,861,447 common and rare variants, 18,025 genes, of which, 5,288 genes were mapped to 259 pathways downloaded from the KEGG database) and 11 phenotypes: high density lipoprotein cholesterol (HDL), low density lipoprotein cholesterol (LDL), triglyceride (Trig) and total cholesterol (TotChol), fast glucose, systolic blood pressure (SBP), diastolic blood pressure, body mass index (BMI) , fastinsulin, Fibrinogen, and platelet count (PLATELET) (no missing phenotype data). Inverse rank normal transformation of the phenotypes was used in the analysis.

The analysis consisted of two stages. At the first stage, the sparse functional SEMs were applied to each of the 259 pathways and 11 phenotypes to infer genotype-phenotype networks. The remaining 12,737 genes which were not mapped to KEGG pathways were divided into 100 groups according to the order of chromosomes. Again, the sparse functional SEMs were applied to each group of genes and 11 phenotypes. We identified 1,789 genes with P-values for testing path coefficients < 0.05 from the analysis at the first stage. To dissect pleiotropic genetic structure, at the second stage, we select 142 genes that were connected with more than one phenotype for further analysis. The sparse functional SEMs were applied to the selected 142 genes and 11 phenotypes to infer genotype-phenotype networks. To improve the accuracy of estimation, a stability selection procedure was used to infer the structure of the network. In other words, we randomly resampled data and estimated the genotype-phenotype networks 100 times. We only selected arrows when their P-values for testing the path-coefficients were less than 0.05 and they were present in the estimated network more than 80 times, i.e., the probability for each arrow to be selected was more than 0.8.  We identified a genotype-phenotype network with 137 genes directly connected to 11 phenotypes and 341 edges. 114 genes out of 137 genes showed pleiotropic genetic effects. The results were presented in Figure 5. We observed that the most causal relationships among phenotypes had P-values $< 10^{-7}$ and stability ~1. This showed that the inference about phenotype sub−network is highly reliable. We also observed that large proportion of the edges in the phenotype network had two directions. This demonstrated that the SEMs had limitations for inferring causal networks.

We observed that *PIK3R5* directly affected 7 phenotypes, *HS1BP3* directly affected 5 phenotypes, 11 genes directly affected 4 phenotypes, and 33 genes directly affected 3 phenotypes and remaining 102 genes directly affected 2 phenotypes. To assess the roles of path analysis in detecting genetic pleiotropic effects, we presented Table 1 that summarizes the P-values of 3 genes that affecting more than 4 phenotypes for path coefficients, the marginal effects of single and multiple traits (simple regression and multiple regression), and the minimum of P-values derived from principal component analysis (PCA) based regression. Table 1 showed that the most P-values for path coefficient were less than that for the marginal effect of corresponding single trait. Estimation of the marginal genetic effect of single trait only explores information of the target trait and genetic variants. However, estimation of the path coefficient uses information of all the relevant traits and genetic variants. This implies that path analysis has higher power to detect genetic risk variants than the traditional marginal analysis. From Table 1 we also observed that in general, each gene had at least one path with P-value for path coefficient was less or close to that for marginal effects of multiple relevant traits or their PCA analysis. Table S2 summarizes the results for all the 13 genes that connected to more than 4 phenotypes.

SEMs provide a powerful tool to distinguish four types of effects: direct, indirect, total and marginal (estimated by a simple regression) effects. Table S3 summarized the direct, indirect, total and marginal effects of one variable (phenotype or gene) that was referred to as the causal on another variable (phenotype) that was referred to as outcome in Table S3, for all 11 phenotypes and 137 genes in the Figure 5. Investigating each type of effect allows a more comprehensive understanding of the relationship between variables.

Table S3 listed the total 1,414 pairs of causal relations between variables. We observed 343 (24.3%) pairs of relations with direct effects, 1,283 (90.7%) pairs of relations with indirect effects and 212 (15%) pairs of relations with both direct and indirect effects (Table 2 showed examples of pairs with both direct and indirect effects). This implied that the most effects are indirect effects due to mediation. In the quantitative trait locus (QTL) analysis, we often identify QTL by testing association of the marginal effect with the single trait. The SEMs provide complimentary information about path coefficients. In Table 3 we listed 25 tests in which the P-values for testing path coefficients were smaller than that for testing the marginal effects (coefficient of SRG) and 25 tests in which the P-values for testing marginal effects were smaller than that for testing the path coefficients. This showed that using SEMs for path analysis will discover additional QTLs that may be missed by marginal association analysis. In theory, the total effect of the causal $X$ on outcome $Y$ is equal to the summation of the product of the path coefficients along all possible paths between $X$ and $Y$.[34] In the previous section, the total effect is defined as the coefficient $\beta_{YX.Parent_X}$ of $X$ in the linear regression of $Y$ on $X$ and its parent set. Let $Z = Parent_x$. The total effect $\beta_{YX.Z}$ can be expressed by[34]

$$\beta_{YX.Z} = \beta_{YX} \frac{1 - \frac{\rho_{YZ}\rho_{ZX}}{\rho_{YX}}}{1 - \rho_{XZ}^2}. \qquad (19)$$

Since causal relations between SNPs do not exist, any SNP does not have its parent, i.e., the set $Z = \phi$ is empty. Therefore, for the SNP or gene $X$, we have $\beta_{YX.Z} = \beta_{YX}$. For example, the estimator of the direct effect of gene *MET* on the phenotype SBP was -0.0596. *MET* also had path $MET \rightarrow$ DBP $\rightarrow$ SBP. The indirect effect of *MET* on SBP was $0.0621 \times 0.605 = 0.0376$. Thus, the total effect of *MET* on SBP was -0.022. The marginal effect $\beta_{YX}$ of *MET* on SBP estimated by SRG was -0.0212. The total effect of *MET* on SBP estimated from Figure 5 was close to the marginal effect of *MET* on SBP. This example showed that if the causal relationships among the variables were completely captured by a DAG, the total effect and marginal effect were almost equal. Therefore, in the genotype-phenotype estimation process, we can use the relationship between the total and marginal effects to check whether the causal relationship modeled by a DAG is complete.

Multiple SNPs within a gene jointly have significant genetic effects, but individually each SNP make mild contributions to the phenotype variation. Table 4 listed P-values of 22 SNPs in 7 genes for testing the path coefficients. We observed that single SNP made only a mild contribution to the direct effect, the multiple SNPs made significant contributions to the phenotype variation. This showed that the gene-based genotype-phenotype inference had higher power than the single SNP-based genotype-phenotype inference.

Since the most existing methods for genotype-phenotype network estimation only take a single SNP as a variable (unite of analysis) and cannot take a gene as a unite of analysis, next we illustrate the application of the sparse two stage multivariate SEMs (S2SEM) for inference of genotype-phenotype network using SNPs and compared their results with that of QTL-driven phenotype network method (QTLnet)[22]. The number of SNPs in 137 genes was 5,482. Due to the limitation of the size of the genotype-phenotype network which the sparse multivariate SEMs can estimate, from 137 genes in Figure 5 we selected 45 genes that were reported to be associated with the 11 phenotypes in the analysis or other cardiovascular disease (CVD) related phenotypes in the literature. A total of 1,993 SNPs in the 45 genes (248 common and 1,745 rare SNPs) were included in the analysis.

The gene-based genotype-phenotype network with 55 nodes (11 phenotypes and 44 genes) and 110 edges estimated using the selected 45 genes and the FSEM method was shown in Figure 6. S2SEM can also be used to estimate gene-based genotype-phenotype network. The procedures were as follows. At the first stage, S2SEM method and all 1,993 SNPs were used to estimate the genotype-phenotype network (Figure 7) where a gene was connected to a phenotype if the minimum of P-values for the coefficients of all the paths connecting SNPs within a gene and a phenotype was less than 0.05. At the second stage, we used Bonferroni correction to adjust P-values for multiple tests. In Figure 8, we plotted the estimated genotype-phenotype network with 17 nodes (11 phenotypes and 6 genes) and 22 edges using the selected 45 genes and the gene-based S2SEM method where a gene was connected to a phenotype if the Bonferroni correction adjusted P-values for path coefficients connecting gene and phenotype was less than 0.05. Figures 6 and 8 showed that the gene-based FSEM method can identify much more genes influencing phenotypes than the gene-based S2SEM method.

Next we study the SNP-based genotype-phenotype network estimation using the S2SEM method. In other words, we connected genes to the phenotypes using the minimum of P-values for the coefficients of all the paths that connect SNPs within a gene and a phenotype without Bonferroni correction. Figure 7 plotted the estimated genotype-phenotype network with 42 nodes (11 phenotypes and 31 genes) and 78 edges using S2SEM method and 1,993 SNPs in the 45 genes. The path coefficients and P-values (<0.05) for the path coefficients of the edges connecting the SNPs in the gene to the phenotypes were summarized in Table S4. In Figure S5, we plotted the estimated genotype-phenotype network with 13 nodes (10 connected phenotypes, one isolated phenotype and 2 genes) and 20 edges using QTLnet method. In Table S5, we listed the edges of the estimated network which connect the genes and phenotypes using QTLnet method. While the QTLnet method only identified two genes: *LBP* connected to the phenotypes TRIGS and TOTCHOL, and *DOCK1* connected to the phenotype HDL, the SNP-based and gene-based S2SEM method, respectively, discovered 31 and 6 genes connected to phenotypes. These results showed that all proposed SEM methods including FSEM, gene-based and SNP-based S2SEM methods outperform the QTLnet method.

Similar to the gene-based FSEM method, we observed several remarkable features from these results obtained by the S2SEM method. First, we observed three SNPs that showed pleiotropic genetic effects (rs138251768 in the gene *ADAMTS19* effected SBP and DBP, rs116623954 in the gene *CNIH3* affected FASTINSULIN and FIBRINOGEN, rs13223756 in the gene *MET* affected SBP and DBP). Second, multiple SNPs in the same gene affected the same phenotype. Three SNPs: rs754555, rs754554 and rs754553 in the gene *DFNA5* jointly affected BMI, two SNPs: rs11017658 and rs61758438 in the gene *DOCK1* jointly affected SBP. Third, the pleiotropic effects of the gene were due to different SNPs. The SNPs: rs564665 and rs141647150 in the gene *DAB1* affected phenotypes DBP and FASTGLUCOSE, respectively; the SNPs: rs376043577 and rs3731878 in the gene *IHH* affected BMI and PLATELET COUNT, respectively; SNPs rs2232585 and rs2232605 in the gene *LBP* affected FIBRINOGEN and PLATELET COUNT, respectively. SNPs rs2305610, rs372123385 and rs17027957 in the gene *OSBPL10* affected BMI, DBP and FIBRINOGEN, respectively; three SNPs: rs144082896, rs140962261 and rs11547635 in the gene *SYN3* affected TRIGS, SBP and FASTINSULIN, respectively. You can find more examples from Tables S4 and S5. Due to space limitation, they are omitted here.

In summary, we jointly estimated genetic architecture and phenotype network with 137 genes that were significantly connected to phenotypes. A total of 45 genes out of 137 genes were reported to be associated with 11 phenotypes or CVD related phenotypes, Table S6 summarized the results of the reported 45 genes and their associated phenotypes. For the reported phenotypes, 6 phenotypes are from the analyzed 11 phenotypes. According to Figure 5, Gene *SMC2* was connected with phenotypes: BMI, HDL and FIBRINOGEN. It was reported associated with HDL and BMI,[43,44] and also related with respiratory function and Echocardiography.[45,46] Gene *RNF157* was connected with HDL, and it was reported associated with blood pressure[47] and HDL.[43] The other pairs of association for these 6 phenotypes were found through indirect paths from Figure 5. For example, gene *DAB1*, *DFNA5* and *DOCK1* were reported associated with LDL[44], and there are indirect path from these genes to LDL according to Figure 5. From these results we can summarized that our gene-based functional SEMs (FSEM) has a rather high power to detect genetic pleiotropic effects, and it also provide a tool to decompose the effects into direct and indirect effects.

## Discussion

Alternative to the standard marginal models for genetic association analysis of multiple correlated phenotypes, we have developed sparse SEMs and sparse FSEMs as a statistical framework for joint analysis of genetic architecture and causal phenotype network, which may emerge as a new generation of genetic analysis of multiple phenotypes exploring the causal network structures of the phenotypes. To facilitate using SEMs as a new paradigm for genetic analysis of multiple phenotypes, several issues have been addressed in this paper.

The first issue is to develop a unified framework for joint analysis of genetic architecture and causal phenotype network with both GWAS and the NGS data. The traditional multivariate

SEMs can be applied to infer genotype-phenotype network with common variants, but are difficult to deal with rare variants. To overcome this limitation, we extend the multivariate SEMs to functional SEMs where exogenous genotype profiles across a genomic region or a gene are represented as a function of the genomic position for genetic analysis of multiple quantitative traits. In other words, we extend the variant-based genotype-phenotype network analysis to gene-based genotype-phenotype network analysis.

The second issue is how to develop statistical methods for jointly inferring genetic architecture and casual phenotype network structure. There is increasing consensus that the structure of the network in nature is sparse. However, the traditional estimation methods for the SEMs do not take the sparsity presented in the network into account. To solve this problem, we developed sparse SEMs and sparse functional SEMs to automatically incorporate the sparse condition into the estimation process. The widely used estimation method for the SEMs is the maximum likelihood method. However, the penalized maximum likelihood method and coordinate descent algorithms are not scalable to SEMs of high dimension. To overcome this limitation, we develop the ADMM-based sparse two-stage least square estimation method for the structure and parameter estimation of the SEMs. Our experience showed that the newly developed ADMM-based sparse two-stage least square estimation methods can infer networks with hundreds of nodes.

The third issue is the true structure discovery. An essential problem for the genotype-phenotype network analysis is to accurately estimate the network structure. By large scale simulations we showed that the true network structure can be accurately recovered with high probability. We also compared the performance of the sparse two-stage least square estimate methods with the QTLnet method. We demonstrated that for all the three cases (common, rare and both common and rare variants) our sparse two-stage SEMs (S2SEM) outperformed QTLnet method. Since the gene-based version of QTLnet method has not been developed we only compared the power and false discovery rates of the variant-based SEMs and gene-based functional SEMs. We found that for all spectrums of allele frequencies (common, rare and both common and rare variants) the gene-based functional SEMs substantially outperformed the variant-based multivariate SEMs.

The fourth issue is how to distinguish four types of effects: direct, indirect, total and marginal effects. The current paradigm for genetic association analysis of multiple phenotypes is genetic marginal analysis in which the effects of the genetic variants on the phenotypes are estimated by regressing phenotypes on the genetic variants. This paradigm is unable to unravel the structure of the genotype-phenotype network and to estimate direct, indirect and total effects of the genetic variants on the phenotypes. The direct, indirect and total genetic effects provide valuable information for dissecting genetic structure of complex traits. We developed sparse SEMs and FSEMs as a causal inference tool to estimate direct, indirect and total genetic effects in addition to estimating marginal genetic effects. We observed that the most effects were indirect effects due to mediation. In traditional QTL analysis, we often identify QTL by testing association of the marginal effect with the single trait. The FSEMs and SEMs provide complimentary information about path coefficients. Interestingly, we found that many P-values for testing path coefficients were smaller than that for testing the marginal effects. This demonstrated that only using marginal association analysis we might miss identification of many significant QTLs.

The fifth issue is how to solve the large genotype-phenotype networks with up to hundreds of nodes or genes. A key to the large network inference is computation efficiency of the algorithms. Two strategies were employed to solve this problem. The first strategy was to reduce the dimension of data using functional data analysis. We first expand the genotype profiles in a genomic region (gene) in terms of orthonormal eigenfunctions. Genetic information across all variants in the genomic region including all single variant variation and their linkage disequilibrium is compressed into functional principal component scores. We use genetic information compressed into functional principal component scores to infer genotype-phenotype networks. The second strategy is to use ADMM algorithms to optimally solve the sparse SEM problem. The widely used algorithms for sparse SEMs are coordinate descent algorithms borrowed from the lasso originally designed for the sparse linear regression. The ADMM algorithms are parallel and efficient. Their convergence rates are fast. The ADMM algorithms allow inferring networks with hundreds or even thousands of nodes.

Major limitation of the SEMs for joint inference of genetic architecture and causal phenotype networks is the presence of two directions associated with one edge in the estimated network, which leads to a cyclic graph. To remove the cycles from the graph we need to strictly enforce the global constraint that the graph structure has to be acyclic. Such problems are often casted into a combinatorial optimization problem. We rank graph structures via a scoring metric that measure how well the DAG models fit the data. Combinatorial optimization algorithms are then used to search the optimal DAG with the best score.[48]

Although their application to genome-wide genotype-phenotype network construction is difficult due to computational limitations, the SEMs are suitable to the phenome-wide association studies where starting phenomics, defined as the unbiased study of a large number of phenotypes in a population. We study the complex networks between multiple expressed phenotypes and genetic variants. Since the number of genetic variants in the phenome-wide association is quite limited and hence the size of the genotype-phenotype network is limited, the required computational time of construction of genotype-phenotype networks using SEMs is in the range the current computer system can reach. Advances in biosensors and sequencing technologies generate large amounts of phenotype and genetic data. SEMs and causal inference may emerge as a new paradigm of genetic studies of complex traits. The main purpose of this paper is to stimulate discussions about what are the optimal strategies to facilitate the development of a new generation of genetic analysis. We hope that our results will greatly increase the confidence in joint inference of genetic architecture and causal phenotype networks.

## Conflict of Interest

The authors declare no conflict of interest.

## Acknowledgments

Mr. Xiong was supported by Grant 1R01AR057120–01 and 1R01HL106034-01, from the National Institutes of Health and NHLBI. Ms. Wang and Mr. Jin were supported by Grant 31330038 from the National Science Foundation of China and grant B13016 from the Programme of Introducing Talents of Discipline to Universities (111 Project).

The authors wish to acknowledge the support of the National Heart, Lung, and Blood Institute (NHLBI) and the contributions of the research institutions, study investigators, field staff and study participants in creating this resource for biomedical research. Funding for GO ESP was provided by NHLBI grants RC2 HL-103010 (HeartGO), RC2 HL-102923 (LungGO) and RC2 HL-102924 (WHISP). The exome sequencing was performed through NHLBI grants RC2 HL-102925 (BroadGO) and RC2 HL-102926 (SeattleGO).

**Supplementary Information**

Supplementary information is available at European Journal of Human Genetics' website.

**Titles and Legends to Figures**

**Figure 1.** The power of two methods: S2SEM and QTLnet for detecting the structure of the genotype (common variants, rare variants and both common and rare variants) – phenotype network as a function of sample size, assuming the network sparsity level $= 0.008$.

**Figure 2.** The false discovery rates of two methods: S2SEM and QTLnet for detecting the structure of the genotype (common variants, rare variants and both common and rare variants) – phenotype network as a function of sample size, assuming the network sparsity level $= 0.008$.

**Figure 3.** The power curves of the S2SEM with the SNP-based and gene-based methods for detecting the structure of the genotype (common variants, rare variants and both common and rare variants) – phenotype network as a function of sample size, assuming the network sparsity level $= 0.008$.

**Figure 4.** The false discovery rates of the S2SEM with the SNP-based and gene-based methods for detecting the structure of the genotype (common variants, rare variants and both common and rare variants) – phenotype network as a function of sample size, assuming the network sparsity level $= 0.008$.

**Figure 5.** A genotype-phenotype network consisted of 137 genes and 11 phenotypes, 114 genes of them showed pleiotropic genetic effect. The nodes in yellow color represented the phenotypes, the nodes in light red color represented genes influencing phenotype variation, the nodes in the red color represented genes that are reported to be associated with 11 phenotypes or cardiovascular diseases phenotypes, the black arrows indicated the causal relations between phenotypes, the blue arrows indicted the contribution of the gene to one phenotype.

**Figure 6.** A genotype-phenotype network consisted of 44 genes and 11 phenotypes constructed using FSEM from 45 genes. Nodes and edges are the same as described in Figure 5.

**Figure 7.** A genotype-phenotype network consisted of 31 genes and 11 phenotypes constructed using SNP-based S2SEM method from 1993 SNPs of 45 genes. Nodes and edges are the same as described in Figure 5.

**Figure 8.** A genotype-phenotype network consisted of 6 genes and 11 phenotypes constructed using the gene-based S2SEM method from 1993 SNPs of 45 genes where a gene was connected to a phenotype if the Bonferroni correction adjusted P-values for path coefficients connecting gene and phenotype was less than 0.05. Nodes and edges are the same as described in Figure 5.

## Appendix A

## Alternating Direction Method of Multipliers for Sparse SEMs

The optimization problem (7) can be further reduced to

$$\begin{aligned} &\min \quad f(\Delta_i) + \lambda \| Z_i \|_1 \\ &\text{subject to} \quad \Delta_i - Z_i = 0. \end{aligned} \quad \text{(A1)}$$

To solve the optimization problem (A1), we form the augmented Lagrangian

$$L_\rho(\Delta_{i,} Z_i, \mu) = f(\Delta_i) + \lambda \| Z_i \|_1 + \mu^T (\Delta_i - Z_i) + \frac{\rho}{2} \| \Delta_i - Z_i \|_2^2 . \quad \text{(A2)}$$

The alternating direction method of multipliers (ADMM) consists of the iterations:

$$\Delta_i^{(k+1)} :== \arg\min_{\Delta_i} L_\rho(\Delta_i, Z_i^{(k)}, \mu^{(k)}) \quad \text{(A3)}$$

$$Z_i^{(k+1)} :== \arg\min_{Z_i} L_\rho(\Delta_i^{(k+1),}, Z_i, \mu^{(k)}) \quad \text{(A4)}$$

$$\mu^{(k+1)} :== \mu^{(k+1)} + \rho(\Delta_i^{(k+1)} - Z_i^{(k+1)}), \quad \text{(A5)}$$

where $\rho > 0$. Let $u = \frac{\mu}{\rho}$. Equations (A1-A3) can be reduced to

$$\Delta_i^{(k+1)} :== \arg\min_{\Delta_i} \left( f(\Delta_i) + \frac{\rho}{2} \| \Delta_i - Z_i^{(k)} + u^{(k)} \|_2^2 \right) \quad \text{(A6)}$$

$$Z_i^{(k+1)} :== \arg\min_{Z_i} \left( \lambda \| Z_i \|_1 + \frac{\rho}{2} \| \Delta_i^{(k+1)} - Z_i + u^{(k)} \|_2^2 \right) \quad \text{(A7)}$$

$$u^{(k+)} :== u^{(k)} + (\Delta_i^{(k+1)} - Z_i^{(k+1)}). \quad \text{(A8)}$$

Solving minimization problem (A6), we obtain

$$\Delta_i^{(k+1)} = [W_i^T X (X^T X)^{-1} X^T W_i + \rho I]^{-1} [W_i^T X (X^T X)^{-1} X^T y_i + \rho(Z_i^k - u^k)],$$

which can be reduced to

$$\Delta_i^{(k+1)} = [\frac{1}{\rho} I - \frac{1}{\rho} W_i^T X (\rho X^T X + X^T W_i W_i^T X)^{-1} X^T W_i][W_i^T X (X^T X)^{-1} X^T y_i + \rho(Z_i^k - u^k)] \quad \text{(A9)}$$

The optimization problem (A7) is non-differentiable. Although the first term in (A7) is not differentiable, we still can obtain a simple closed-form solution to the problem (A7) using subdiffenrential calculus.[37] Let $\Gamma_j$ be a generalized derivative of the $j$ -th component $Z_i^j$ of the vector $Z_i$ and $\Gamma = [\Gamma_1, ..., \Gamma_{M+K-1}]^T$ where

$$\Gamma_j = \begin{cases} 1 & Z_i^j > 0 \\ [-1,1] & Z_i^j = 0 \\ -1 & Z_i^j < 0 \end{cases}$$

Then, we have

$$\frac{\lambda}{\rho}\Gamma + Z_i = \Delta_i^{k+1} + u^k,$$

which implies that

$$Z_i^{(k+1)} = \mathrm{sgn}(\Delta_i^{k+1} + u^k)(|\Delta_i^{k+1} + u^k| - \frac{\lambda}{\rho})_+, \qquad \text{(A10)}$$

where

$$|x|_+ = \begin{cases} x & x \geq 0 \\ 0 & x < 0. \end{cases}$$

## Appendix B

## Estimation of Parameters in the Sparse Structural Functional Equation Models for the Genotype-Phenotype Networks

Assume that the sparse SFEMs are given by

$$\begin{aligned}
&y_1\gamma_{11} + y_2\gamma_{21} + ... + y_M\gamma_{M1} + \int_{T_1} x_1(t)\beta_{11}(t)dt + ... + \int_{T_k} x_k(t)\beta_{k1}(t)dt + e_1 = 0 \\
&y_1\gamma_{12} + y_2\gamma_{22} + ... + y_M\gamma_{M2} + \int_{T_1} x_1(t)\beta_{12}(t)dt + ... + \int_{T_k} x_k(t)\beta_{k2}(t)dt + e_2 = 0 \\
&\vdots \qquad\qquad\qquad\qquad \vdots \qquad\qquad\qquad\qquad \vdots \\
&y_1\gamma_{1M} + y_2\gamma_{2M} + ... + y_M\gamma_{MM} + \int_{T_1} x_1(t)\beta_{1M}(t)dt + ... + \int_{T_k} x_k(t)\beta_{kM}(t)dt + e_M = 0
\end{aligned} \qquad \text{(B1)}$$

For each genomic region or gene, we use functional principal component analysis to calculate principal component function.[14] We expand $x_{ij}(t),\ j = 1,2,...,k$ in each genomic region in terms of orthogonal principal component functions:

$$x_{ij}(t) = \sum_{l=1}^{L_j} \eta_{ijl}\phi_{jl}(t),\ j = 1,...,k, \qquad \text{(B2)}$$

where $\phi_{jl}(t),\ j = 1,...,k, l = 1,...,L_j$ are the $l$ -th principal component function in the $j$ -th genomic region and $\eta_{ijl}$ are the functional principal component scores of the $i$ -th individual. Using the functional principal component expansion of $x_{ij}(t)$, we obtain

$$\int_T x_{ij}(t)\beta_{jm}(t)dt = \int_T \sum_{l=1}^{L_j} \eta_{ijl}\phi_{jl}(t)\beta_{jm}(t)dt = \sum_{l=1}^{L_j} \eta_{ijl}b_{jlm}, i = 1,...,n,\ j = 1,...,k, m = 1,...,M\ . \qquad \text{(B3)}$$

Let $x_j(t) = [x_{1j}(t),...,x_{nj}(t)]^T, \eta_{jl} = [\eta_{1jl},...,\eta_{njl}]^T$. Substituting equation (B3) into equation (B1), we obtain

$$
\begin{aligned}
&y_1 r_{11} + y_2 r_{21} + \cdots + y_M r_{M1} + \sum_{l=1}^{L_1} \eta_{1l} b_{1l1} + \cdots + \sum_{l=1}^{L_k} \eta_{kl} b_{kl1} + e_1 = 0 \\
&\vdots \qquad\qquad\qquad\qquad \vdots \qquad\qquad\qquad\qquad \vdots \\
&y_1 r_{1M} + y_2 r_{2M} + \cdots + y_M r_{MM} + \sum_{l=1}^{L_1} \eta_{1l} b_{1lM} + \cdots + \sum_{l=1}^{L_k} \eta_{kl} b_{klM} + e_M = 0
\end{aligned} \tag{B4}
$$

Let $\eta = [\eta_{11},...,\eta_{1L_1,...,}\eta_{k1},...,\eta_{kL_k}]$, $B = \begin{bmatrix} b_{111} & \cdots & b_{11M} \\ \vdots & \vdots & \vdots \\ b_{1L_11} & \cdots & b_{1L_1M} \\ \vdots & \vdots & \vdots \\ b_{k11} & \cdots & b_{k1M} \\ \vdots & \vdots & \vdots \\ b_{kL_k1} & \vdots & b_{kL_kM} \end{bmatrix}$ and $Y, \Gamma$ and $E$ be defined as before.

In matrix form, equation (B4) can be rewritten as

$$
Y\Gamma + \eta B + E = 0, \tag{B5}
$$

which has the same form as the equation (2) has.

If we consider only one genomic region or gene, the matrices $\eta$ and $B$ will be reduced to

$$
\eta = [\eta_1,...,\eta_L] \text{ and } B = \begin{bmatrix} b_{11} & \cdots & b_{1M} \\ \vdots & \vdots & \vdots \\ b_{L1} & \cdots & b_{LM} \end{bmatrix}.
$$

If we take functional principal component scores as predictors, the models and algorithms for network structure and parameter estimation will be similar to that discussed in Appendix A. Specifically, the $i$-th equation is given by

$$
Y\Gamma_i + \eta B_i + e_i = 0,
$$

which can be rewritten as

$$y_i = W_i \Delta_i + e_i, \tag{B6}$$

where $W_i = [Y_{-i} \quad \eta], \Delta_i = [\gamma_{-i} \quad B_i]$.

Then, the sparse SFEMs are transformed to

$$\min_{\Delta_i} \quad f(\Delta_i) + \lambda \| \Delta_i \|_1$$
$$\text{where } f(\Delta_i) = (\eta^T y_i - \eta^T W_i \Delta_i)^T (\eta^T \eta)^{-1} (\eta^T y_i - \eta^T W_i \Delta_i). \tag{B7}$$

Finally, ADMM algorithms are given by

Algorithm:

For $i = 1, \ldots, M$

Step 1. Initialization

$$u^0 := 0$$
$$\Delta_i^0 := [W_i^T \eta (\eta^T \eta)^{-1} \eta^T W_i]^{-1} W_i^T \eta (\eta^T \eta)^{-1} \eta^T y_i$$
$$Z_i^0 := \Delta_i^0.$$

Carry out steps 2,3 and 4 until convergence

Step 2.

$$\Delta_i^{(k+1)} := [\frac{1}{\rho} I - \frac{1}{\rho} W_i^T \eta (\rho \eta^T \eta + \eta^T W_i W_i^T \eta)^{-1} \eta^T W_i][W_i^T \eta (\eta^T \eta)^{-1} \eta^T y_i + \rho (Z_i^k - u^k)]$$

Step 3.

$$Z_i^{(k+1)} := \text{sgn}(\Delta_i^{k+1} + u^k)(| \Delta_i^{k+1} + u^k | - \frac{\lambda}{\rho})_+.$$

Step 4.

$$u^{(k+)} :== u^{(k)} + (\Delta_i^{(k+1)} - Z_i^{(k+1)}).$$

Table 1. P-values for the path coefficient, mariginal effects of single trait and multiple traits, and minimum of P-values from PCA analysis (example of 3 genes that connected to more than 4 phenotypes).

| Outcome | Causal | Stability | P-value for path coefficient | P-value (Single Trait Marginal ) | Min (P-value) PCA | P-value (Multiple Traits Marginal) |
|---|---|---|---|---|---|---|
| | | | | | 1.44E-03 | 1.57E-03 |
| SBP | PIK3R5 | 1 | 8.38E-05 | 9.25E-02 | | |
| DBP | PIK3R5 | 1 | 1.20E-03 | 7.62E-02 | | |
| TRIGS | PIK3R5 | 0.9 | 5.59E-03 | 1.07E-01 | | |
| FASTGLUCOSE | PIK3R5 | 0.9 | 1.31E-02 | 2.01E-01 | | |
| BMI | PIK3R5 | 0.88 | 1.63E-02 | 2.61E-01 | | |
| FASTINSULIN | PIK3R5 | 0.92 | 3.00E-02 | 7.94E-01 | | |
| HDL | PIK3R5 | 0.81 | 3.84E-02 | 2.39E-01 | | |
| | | | | | 2.17E-04 | 2.43E-04 |
| DBP | HS1BP3 | 0.96 | 1.21E-04 | 1.58E-01 | | |
| HDL | HS1BP3 | 0.98 | 6.78E-04 | 8.15E-04 | | |
| SBP | HS1BP3 | 0.87 | 6.08E-03 | 9.10E-01 | | |
| BMI | HS1BP3 | 0.94 | 1.71E-02 | 1.91E-01 | | |
| FASTGLUCOSE | HS1BP3 | 0.91 | 1.99E-02 | 5.95E-02 | | |
| | | | | | 4.79E-04 | 5.24E-04 |
| DBP | ABCC9 | 1 | 5.30E-06 | 6.49E-02 | | |
| SBP | ABCC9 | 0.94 | 2.56E-04 | 4.63E-01 | | |
| FASTINSULIN | ABCC9 | 0.95 | 2.71E-03 | 2.10E-03 | | |
| FIBRINOGEN | ABCC9 | 0.96 | 5.52E-03 | 7.71E-02 | | |

Table 2. An example of 20 pairs of variables that had both direct and indirect effects.

| Outcome | Causal | Direct Effect | Indirect Effect | Total Effect | Marginal Effect |
|---|---|---|---|---|---|
| BMI | ATP6V1G3 | -0.5701 | 0.0107 | -0.5594 | -0.5360 |
| BMI | C12orf77 | -1.3704 | -0.2745 | -1.6449 | -1.7371 |
| BMI | EPHB2 | -0.2627 | 0.0544 | -0.2083 | -0.2017 |
| BMI | PIK3R5 | 0.0631 | -0.0119 | 0.0512 | 0.0541 |
| BMI | RPS21 | 0.7630 | 0.0123 | 0.7754 | 0.8282 |
| DBP | CHUK | 0.0553 | -0.0355 | 0.0199 | 0.0213 |
| FASTGLUCOSE | C11orf49 | 0.6186 | 0.0146 | 0.6332 | 0.5846 |
| FASTGLUCOSE | C12orf77 | -1.2863 | -0.2366 | -1.5228 | -1.5121 |
| FASTGLUCOSE | PIK3R5 | 0.0674 | -0.0133 | 0.0541 | 0.0530 |
| FASTGLUCOSE | TAF5L | 0.2911 | 0.0196 | 0.3107 | 0.2942 |
| FASTINSULIN | SFMBT2 | 0.2874 | -0.0898 | 0.1975 | 0.1930 |
| FIBRINOGEN | FAM120AOS | -0.1550 | 0.0033 | -0.1516 | -0.1597 |
| HDL | CREBBP | 0.1168 | -0.0176 | 0.0992 | 0.1097 |
| HDL | ITPR2 | 0.0946 | -0.0542 | 0.0404 | 0.0434 |
| LDL | TOTCHOL | 0.9458 | 0.0161 | 0.9619 | 0.9398 |
| SBP | AQPEP | -0.0425 | 0.0282 | -0.0143 | -0.0156 |
| SBP | CHUK | -0.0595 | 0.0335 | -0.0261 | -0.0272 |
| SBP | MET | -0.0596 | 0.0376 | -0.0220 | -0.0212 |

Table 3. 25 pairs of P-values for testing path coefficients and marginal effects, respectively.

| | | P-value | | | | P-value | |
|---|---|---|---|---|---|---|---|
| Outcome | Causal | Path Coeff | Marginal Effect | Outcome | Causal | Path Coeff | Marginal Effect |
| DBP | ABCC9 | 5.30E-06 | 6.49E-02 | HDL | ITFG2 | 4.42E-04 | 1.98E-05 |
| DBP | TRPM4 | 9.51E-06 | 7.24E-02 | TRIGS | ST3GAL3 | 2.21E-04 | 1.16E-04 |
| BMI | POLR2F | 1.14E-05 | 3.93E-04 | PLATELET | CRTAP | 3.79E-04 | 1.58E-04 |
| PLATELET | TRMT61B | 1.17E-05 | 1.02E-04 | HDL | ASCC2 | 5.58E-04 | 1.99E-04 |
| SBP | CHUK | 1.44E-05 | 1.76E-01 | FASTGLUCOSE | PTH1R | 5.72E-03 | 2.00E-04 |
| DBP | QARS | 2.00E-05 | 8.82E-03 | PLATELET | PDE1B | 4.61E-04 | 3.02E-04 |
| PLATELET | TCOF1 | 2.10E-05 | 1.36E-02 | HDL | HS1BP3 | 6.78E-04 | 3.56E-04 |
| FIBRINOGEN | SMC2 | 2.33E-05 | 1.06E-03 | TRIGS | ITFG2 | 7.32E-03 | 3.81E-04 |
| HDL | DRGX | 3.20E-05 | 3.22E-04 | HDL | NT5E | 3.26E-02 | 5.86E-04 |
| DBP | CHUK | 3.33E-05 | 2.88E-01 | HDL | EPHB2 | 4.30E-03 | 6.33E-04 |
| FIBRINOGEN | C11orf49 | 4.58E-05 | 5.10E-05 | FIBRINOGEN | CEP170B | 4.71E-03 | 7.04E-04 |
| LDL | LARP7 | 4.58E-05 | 3.33E-01 | DBP | PIDD | 1.51E-03 | 7.53E-04 |
| LDL | CD7 | 5.16E-05 | 7.68E-02 | HDL | HBEGF | 3.43E-03 | 8.44E-04 |
| FIBRINOGEN | CRK | 8.66E-05 | 1.45E-03 | FASTGLUCOSE | RNF122 | 4.29E-02 | 9.17E-04 |
| PLATELET | MYLK | 9.80E-05 | 1.14E-03 | DBP | AC053503.11 | 5.62E-03 | 1.08E-03 |
| SBP | STX3 | 1.07E-04 | 1.60E-03 | FIBRINOGEN | CNIH3 | 1.07E-02 | 1.13E-03 |
| FASTINSULIN | MPO | 1.09E-04 | 3.22E-03 | BMI | NFIC | 1.74E-03 | 1.15E-03 |
| FIBRINOGEN | OSBPL10 | 1.18E-04 | 3.99E-03 | FASTGLUCOSE | IHH | 4.66E-02 | 1.17E-03 |
| SBP | ST3GAL3 | 1.56E-04 | 1.99E-03 | FASTGLUCOSE | ARHGAP27 | 1.87E-03 | 1.30E-03 |
| HDL | LARP7 | 1.60E-04 | 3.31E-02 | SBP | SLC18A2 | 1.42E-02 | 1.44E-03 |
| HDL | SOX13 | 1.60E-04 | 6.78E-03 | BMI | QKI | 1.93E-03 | 1.61E-03 |
| PLATELET | SH3TC1 | 2.18E-04 | 1.25E-02 | FASTGLUCOSE | SLC38A1 | 1.44E-02 | 1.65E-03 |
| SBP | ABCC9 | 2.56E-04 | 4.63E-01 | SBP | ZNF740 | 4.68E-03 | 1.81E-03 |
| BMI | DCDC2B | 2.64E-04 | 8.70E-02 | TRIGS | ADAMTS19 | 2.31E-02 | 1.82E-03 |
| SBP | TRPM4 | 2.71E-04 | 2.19E-01 | PLATELET | GAK | 1.54E-02 | 2.02E-03 |

Table 4. P-values of 22 SNPs in 7 genes for testing path coefficients.

| Phenotype | | | | P-value Testing Path Coef | |
|---|---|---|---|---|---|
| | Gene | Chr | SNP Position | Gene | SNP |
| FASTGLUCOSE | SEMA3B | 3 | 50310922 | 5.98E-06 | 6.41E-05 |
| FASTGLUCOSE | DNAJC16 | 1 | 15873386 | 1.09E-05 | 5.62E-03 |
| FASTGLUCOSE | DNAJC16 | 1 | 15874961 | | 3.61E-03 |
| FASTGLUCOSE | DNAJC16 | 1 | 15905501 | | 3.94E-02 |
| DBP | OBSCN | 1 | 228404668 | 1.48E-05 | 8.81E-02 |
| DBP | OBSCN | 1 | 228461187 | | 9.96E-02 |
| DBP | OBSCN | 1 | 228482028 | | 9.68E-02 |
| DBP | OBSCN | 1 | 228496066 | | 8.61E-02 |
| DBP | OBSCN | 1 | 228503711 | | 6.52E-04 |
| DBP | OBSCN | 1 | 228565208 | | 2.66E-03 |
| DBP | OBSCN | 1 | 228565445 | | 8.28E-02 |
| HDL | SOX13 | 1 | 204085609 | 4.31E-05 | 2.45E-02 |
| HDL | SOX13 | 1 | 204092129 | | 9.14E-04 |
| HDL | SOX13 | 1 | 204094963 | | 9.40E-02 |
| HDL | SOX13 | 1 | 204095220 | | 3.56E-02 |
| HDL | SOX13 | 1 | 204095280 | | 3.79E-02 |
| HDL | SRRM5 | 19 | 44099538 | 4.74E-05 | 6.37E-02 |
| HDL | SRRM5 | 19 | 44111890 | | 2.21E-05 |
| FIBRINOGEN | SLC45A4 | 8 | 142225990 | 1.63E-04 | 3.97E-02 |
| FIBRINOGEN | SLC45A4 | 8 | 142226108 | | 1.61E-02 |
| FIBRINOGEN | SLC45A4 | 8 | 142228909 | | 8.87E-04 |
| FIBRINOGEN | LHFPL2 | 5 | 77784738 | 5.95E-06 | 4.21E-06 |

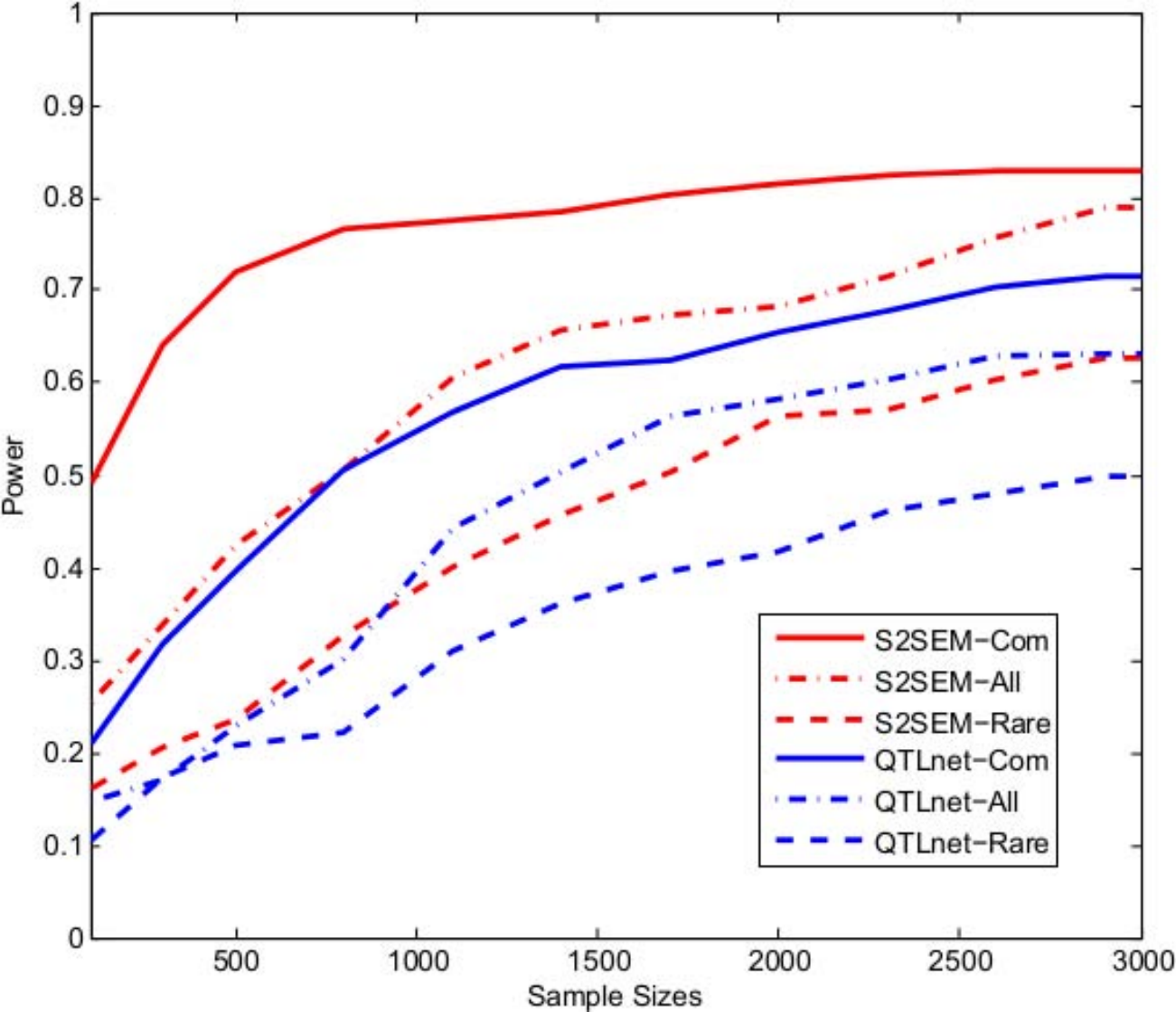

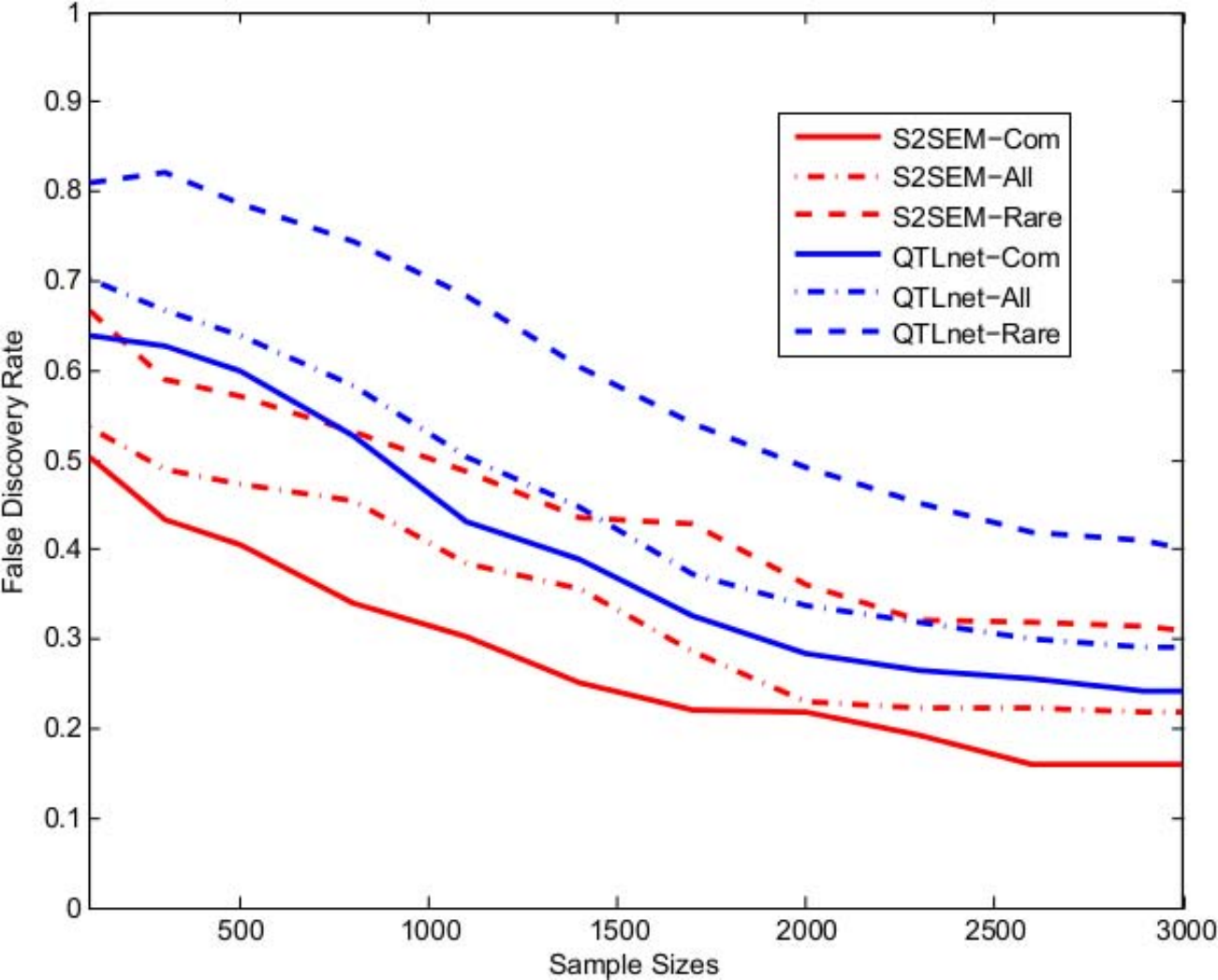

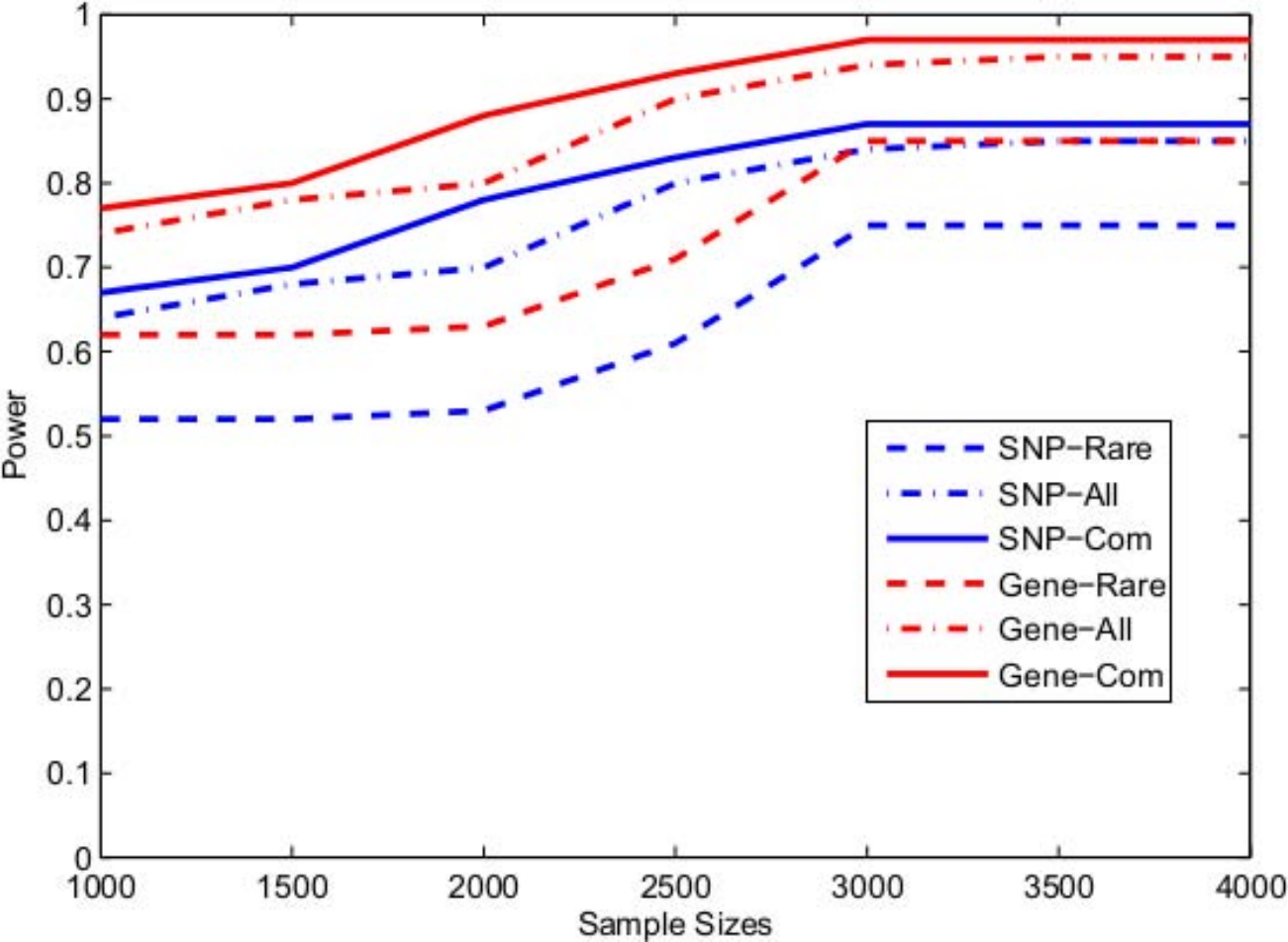

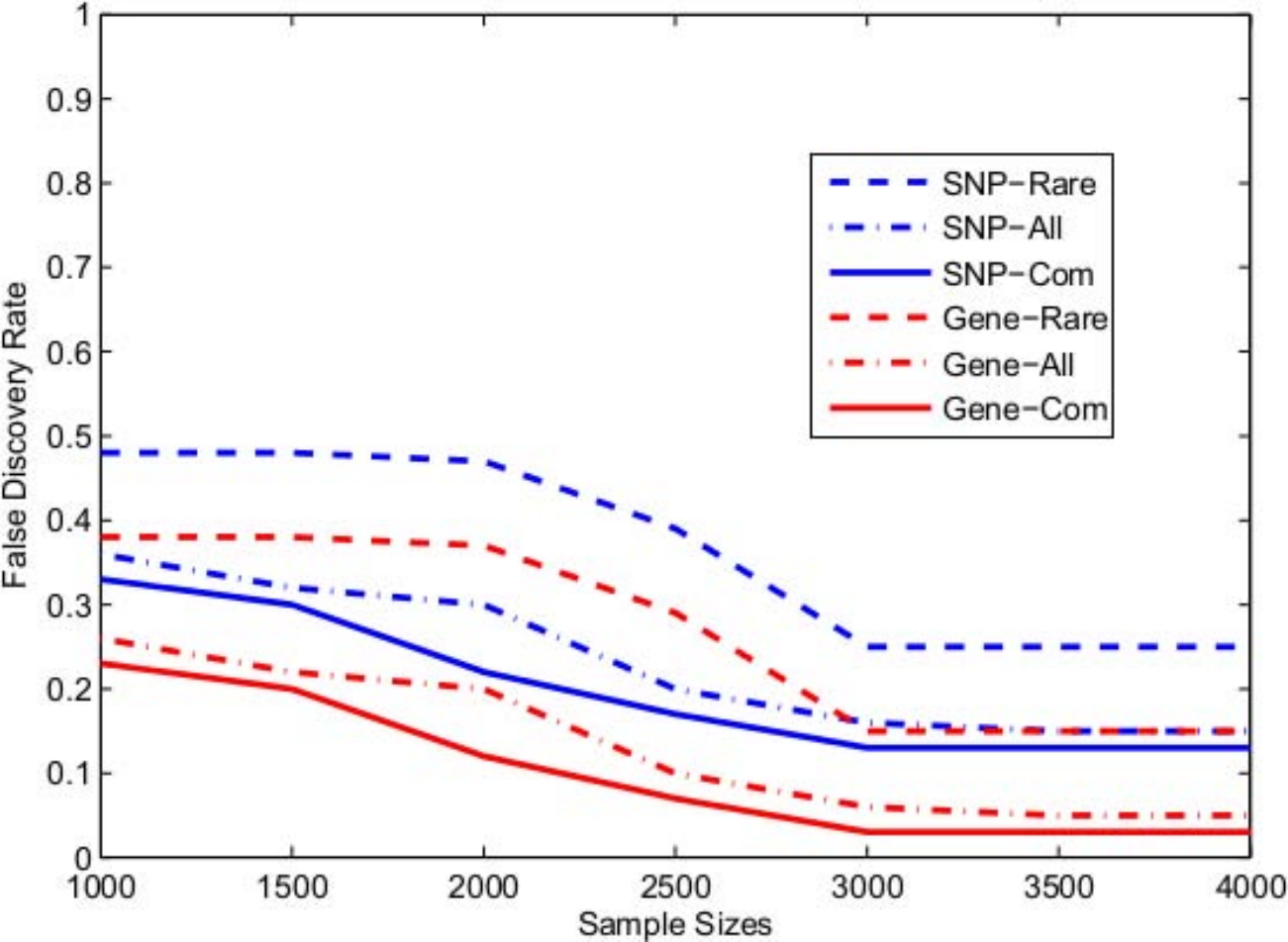